\documentclass[10pt,twocolumn,letterpaper]{article}
\usepackage[pagenumbers]{cvpr}

\usepackage{wrapfig}
\usepackage{multirow}
\usepackage{algorithm}
\usepackage{algpseudocode}
\usepackage{multicol}
\usepackage{amsfonts}
\usepackage{colortbl}

\definecolor{cvprblue}{rgb}{0.21,0.49,0.74}
\usepackage[pagebackref,breaklinks,colorlinks,allcolors=cvprblue]{hyperref}

\definecolor{ggray}{rgb}{0.90, 0.90, 0.98}

\title{CLARE: Scalable Class-Incremental Continual Learning via a \\ Sparsity-Based Framework}

\author{
Yunxiang Fu\textsuperscript{1} \quad Meng Lou\textsuperscript{1} \quad
Zicheng Liao\textsuperscript{2} \quad Yizhou Yu\textsuperscript{1}\\
\textsuperscript{1}School of Computing and Data Science, The University of Hong Kong\\
\textsuperscript{2}Hong Kong Generative AI Research and Development Center\\
\texttt{\small yunxiang@connect.hku.hk, loumeng@connect.hku.hk}\\
\texttt{\small zichengliao@gmail.com, yizhouy@acm.org}
}

\def\confName{CVPR}
\def\confYear{2026}

\begin{document}

\maketitle

\begin{abstract}
Continual learning must balance the learning of new knowledge with the retention of previously learned knowledge to incrementally learn tasks from a data stream without catastrophic forgetting.
While leveraging pretrained models has significantly advanced continual learning, existing methods exhibit a scalability bottleneck when trained sequentially on many tasks, suffering from performance degradation due to inter-task interference and loss of plasticity. 
Inspired by evidence that sparse fine-tuning achieves performance comparable to full fine-tuning, this paper presents a novel sparsity-driven continual learning framework. Our continual learning method, termed CLARE, operates in two stages: it first identifies a sparse, task-critical parameter mask via a sparsity-inducing objective, then performs mask-constrained fine-tuning by only optimizing parameters selected by the mask.
This two-stage sparse adapter mechanism enables all tasks to be accumulated within a shared adapter space while reducing destructive interference across tasks.
Extensive experiments demonstrate the scalability of CLARE. On the long task-sequence benchmark Omnibenchmark-1k, CLARE outperforms strong baselines in final accuracy by a large margin, e.g, improving EASE by 4.64\% and 13.34\% after learning 100 tasks, respectively.
\end{abstract}

\section{Introduction}
\label{sec:intro}
The core challenge of continual learning (CL) lies in achieving a balance between the capacity to learn new diverse tasks (\textit{learning plasticity}) and the ability to retain previously learned knowledge without catastrophic forgetting (\textit{memory stability}). 
Class-incremental learning (CIL) stands as one of the most challenging settings in CL, requiring a model to incrementally learn new classes over time without accessing previous task data, while maintaining recognition capacity on all seen classes.
Traditional CIL methods can often be categorized into three main paradigms: regularization-based methods~\citep{kirkpatrick2017EWC,li2017lwf}, replay-based methods~\citep{lopez2017gradient}, and optimization-based methods~\citep{farajtabar2020orthogonal}. 
Recent advancements leverage strong pretrained models (PTM) to further improve performance instead of training models from scratch, as pretrained models contain rich prior knowledge learned from large-scale datasets~\cite{dosovitskiy2020vit,zhou2024EASE,wang2022L2P}.
In particular, two prominent directions include leveraging task-specific parameters with a routing mechanism to use the most relevant parameters during inference~\citep{zhou2024EASE,yu2024boosting,gao2025MoAL,wang2022L2P,smith2023coda} and merging task-specific parameters into a single set of parameters for all tasks~\citep{marczak2024magmax,liang2024inflora,Wu2025SDLoRA}. While the former achieves stronger performance, it usually requires the number of stored task-specific parameters to increase linearly with the number of tasks and rely on an accurate routing mechanism to predict the task identity of inputs during inference~\citep{lou2026CARE}. In this work, we focus on the latter paradigm since it is efficient when there are many tasks to be learned, as it does not require storing task-specific parameters.
\par
Although recent PTM-based CIL methods (InfLoRA~\citep{liang2024inflora}, SD-LoRA~\citep{Wu2025SDLoRA}) that do not store task-specific parameters have shown promising performance on short task sequences (e.g., 10 or 20 tasks), scaling these methods to longer task sequences typically means substantial performance sacrifice~\citep{lou2026CARE}. 
This decline stems primarily from an imbalance between interference and plasticity.
As the task sequence lengthens, effective new task learning causes catastrophic forgetting of earlier knowledge as new updates overwrite or conflict with parameters crucial for previous tasks. 
However, effective earlier knowledge preservation restricts a model's capacity to integrate new information, resulting in progressively poorer performance on new tasks.

\begin{figure*}[t]
    \centering
    
    \includegraphics[width=0.43\textwidth]{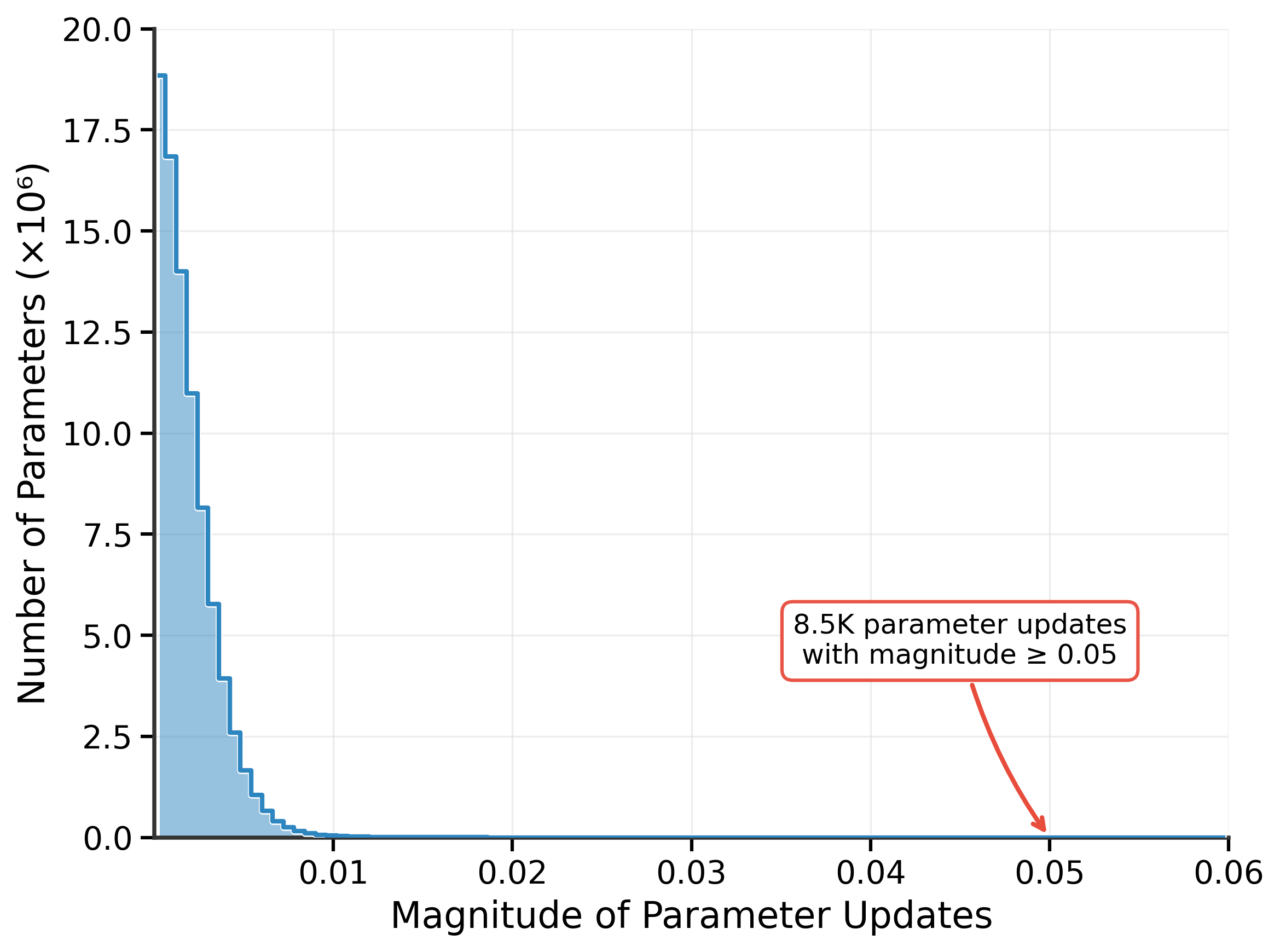}
    \hfill
    \includegraphics[width=0.54\textwidth]{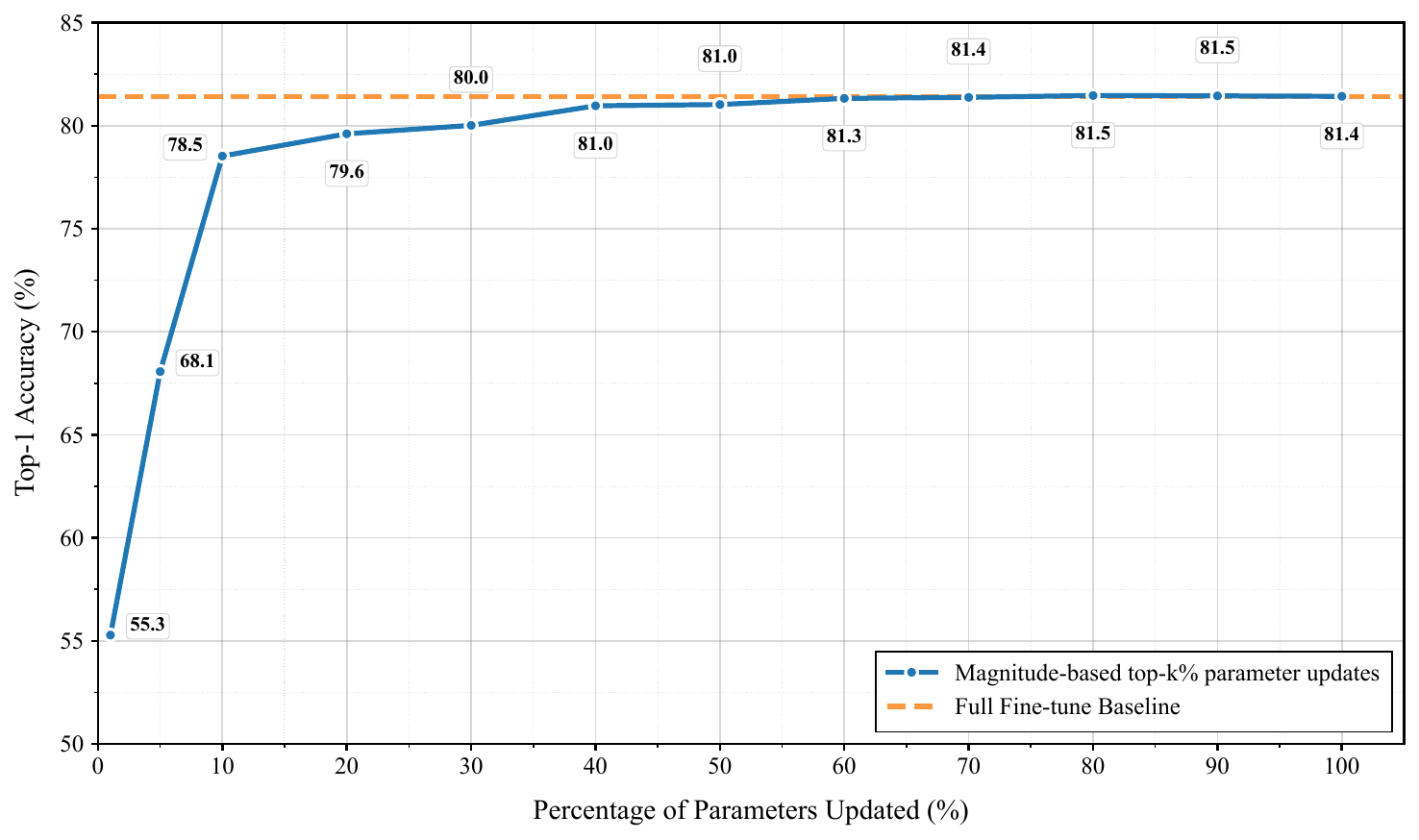}
    \vspace{1em}
    \caption{Sparse parameter update analysis. \textbf{Left:} Long-tail distribution of parameter update magnitudes shows most parameters experience tiny updates ($<$ 0.01), while only 8.5K parameters have updates $\ge$ 0.05. \textbf{Right:} Sparse parameter updates achieve performance close to full fine-tuning on ImageNet-R using ImageNet-1K pretrained ViT-B/16.}
    \label{fig:sparse_analysis}
    \vspace{-1em}
\end{figure*}

In this paper, we hypothesize that strategically learning a small number of parameters for each task can already maintain sufficient plasticity while dramatically reducing the likelihood of destructive interference across tasks.
This hypothesis is supported by existing literature on learning sparse neural networks~\citep{wen2016learning, louizos2017learning,MA2019286} as well as empirical evidence showing that the magnitude of parameter updates during fine-tuning follows a long-tailed distribution (Figure~\ref{fig:sparse_analysis} (left)), with substantial updates being confined to a tiny subset of parameters. More importantly, it is only necessary to update a small proportion of the model parameters to achieve competitive task-specific performance, as illustrated in Figure~\ref{fig:sparse_analysis} (right). 
On the basis of this insight, we propose a sparsity-driven continual learning framework that learns a sparse subset of parameters for each task, enabling effective scaling to extended sequences while balancing the plasticity-stability trade-off in continual learning models.

Our sparsity-driven framework manages parameter allocation across task sequences through a two-stage learning process. Starting from a pretrained base model, we first identify task-critical parameters by optimizing a sparsity-inducing objective, which produces a binary mask identifying the most relevant parameters for the task. We then perform mask-constrained fine-tuning, updating only these relevant parameters while keeping the remainder frozen. This enables the model to achieve promising performance by updating only a sparse subset of the total parameters, thereby facilitating targeted knowledge acquisition with minimal interference and preserved plasticity. In practice, the parameters learned for new tasks are incrementally fused into the base model via simple accumulation for computational efficiency. 
Distinct from InfLoRA~\cite{liang2024inflora} and SD-LoRA~\cite{Wu2025SDLoRA}, our method explicitly learns the subset of parameters that are most important to each task, instead of choosing them based on heuristics such as orthogonal constraints.

We evaluate CLARE through extensive experiments spanning both long and standard task-sequence settings.
On long-sequence benchmarks, CLARE achieves the highest final accuracy of 66.88\% after 100 tasks on OmniBenchmark-1k~\cite{lou2026CARE}, surpassing the strongest adapter-based baseline in final accuracy by 4.64\% and the LoRA-based SD-LoRA~\citep{Wu2025SDLoRA} by over 38\%. On 50-task splits of ImageNet-R and ImageNet-A, CLARE outperforms InfLoRA\citep{liang2024inflora} by 13.9\% and 19.28\% points in final accuracy, respectively. 
On standard class-incremental benchmarks with 10 and 20 tasks, CLARE attains the highest final accuracy across all six evaluated settings, with particularly large margins on datasets with distribution shift: on ImageNet-A and ObjectNet, it improves final accuracy by 5.40\% and 8.40\% points over the strongest baseline, respectively. 
These results demonstrate that capacity-aware sparse adapter learning scales effectively from short to long task sequences and provides a unified solution that does not require task-specific routing or task identity at inference time.
\par
In summary, our contributions are:
\begin{itemize}
\item We propose CLARE, a sparsity-driven adapter-based continual learning framework that learns task-critical sparse adapter masks before mask-constrained task learning.
\item We explicitly learn the task-critical mask using an $L_1$-regularized objective to induce natural sparsity in task-specific parameter updates, allowing each task to discover a compact set of task-critical parameters.
\item We provide extensive evaluations and ablations showing how learned sparsity, two-stage optimization, and adapter capacity affect performance across short, medium, and long task sequences.
\end{itemize}

\section{Related work}

\paragraph{Pretrained Model-Based CIL}
Traditional CIL methods address catastrophic forgetting through regularization constraints~\citep{kirkpatrick2017EWC, li2017lwf}, rehearsal strategies that retain exemplars from previous tasks~\citep{rebuffi2017icarl}, gradient constraints~\citep{lopez2017gradient}, and parameter-isolation mechanisms \citep{mallya2018packnet,aljundi2017expertGate}. Inspired by the development of strong representations learned by pretrained models (PTM) for vision~\citep{dosovitskiy2020vit,radford2021CLIP,fu2025segman,shi2026vision,lou2025overlock,lou2025sparx,lou2025transxnet} and their applicability to different tasks~\citep{li2025semantic,fu2024dreamda,chen2022compound,he2019non,lou2026overcoming}, recent CIL methods increasingly build on frozen or partially tuned pretrained vision transformers \citep{zhou2024EASE,wang2022L2P,lou2026CARE}. 
Prompt-based methods such as L2P~\citep{wang2022L2P}, DualPrompt \citep{wang2022dualprompt}, and CODA-Prompt \citep{smith2023coda} learn task-related prompt tokens and retrieve them at inference. Adapter-based methods further improve plasticity by learning lightweight task-specific modules, as in EASE \citep{zhou2024EASE} and SEMA \citep{wang2025sema}. However, these methods often store task-specific parameters and depend on accurate retrieval or routing during inference, which becomes increasingly challenging as the task sequence grows. In contrast, our work focuses on learning sparse task updates that can be merged into a single task-agnostic model.

\paragraph{Parameter-Efficient CIL}
Another line of PTM-based CIL seeks to avoid task-specific routing by combining task-specific updates into one model for inference. 
Works like InfLoRA~\citep{liang2024inflora}, SD-LoRA~\citep{Wu2025SDLoRA}, and LoDA~\citep{he2026loda} learn task-specific LoRA modules and impose constraints to reduce interference among tasks.
MagMax~\citep{marczak2024magmax} merges task-specific parameter updates into a shared model, using predefined or post-hoc rules such as random pruning or magnitude-based selection. 
There are also conceptually works that focus on LLM tasks, including OA-Adapter~\citep{wan2025OA-adapter}, Share~\citep{kaushik2026shared}, CSBoRA~\citep{liu2026csbora}, OPLoRA~\citep{xiong2026oplora}.
These methods are free of inference-time routing, but do not explicitly learn which parameters should be updated for each task before task learning, leading to suboptimal learning plasticity. In contrast, our method uses an $L_1$-regularized objective to induce naturally sparse task updates, then uses the discovered mask for constrained training and merging.

\paragraph{Sparse and Mask-Based Continual Learning}
Sparsity has also been widely studied in continual learning through subnetwork selection, pruning, and mask learning. Piggyback~\citep{mallya2018piggyback} learns task-specific binary masks over a fixed backbone, while PackNet~\citep{mallya2018packnet} progressively prunes and allocates parameters for new tasks. 
Other sparse or subnetwork-based methods similarly reduce forgetting by assigning different parameter subsets to different tasks~\citep{aljundi2017expertGate,wang2022sparcl,yildirim2024continual,Subnetwork_PGM}. These methods usually maintain task-specific masks or sparse subnetworks and often require task identity or task-specific selection during inference. 
For example, PackNet~\citep{mallya2018packnet} and Piggyback~\citep{mallya2018piggyback} keep a per-task mask/subnetwork and, at test time, select the right one using the task identity or an arg-max over stored masks. They also rely on knowing the number of tasks in advance.

Distinctively, CLARE's core novelty is that sparsity is a training-time allocation mechanism, not an inference-time selection mechanism. 
CLARE keeps no per-task mask at inference. After each task its sparse update is merged into a single shared adapter $\alpha_{\mathrm{shared}}=\alpha_0+\sum_t M_t\!\odot\!(\alpha_t-\alpha_0)$, every input uses that same adapter with all coordinates active, and no task identity or routing is used. The mask $M_t$ only decides which currently-unused coordinates a new task $t$ may modify. A direct consequence is that earlier tasks' coordinates remain active for every later input, so prior knowledge is reused rather than gated. This gives CLARE high learning plasticity (average incremental accuracy), not merely forgetting mitigation.
This also makes sparse parameter selection part of the learning objective rather than a hand-designed or purely post-hoc pruning rule.

\section{Method}

\subsection{Problem Definition}
We study exemplar-free class-incremental learning. The model receives a sequence of image classification tasks $\{\mathcal{D}_1, \ldots, \mathcal{D}_T\}$. Task $t$ contains training samples $\mathcal{D}_t=\{(x_i,y_i)\}_{i=1}^{n_t}$ whose labels belong to a new class set $\mathcal{C}_t$. The class sets are disjoint for different tasks, so $\mathcal{C}_i \cap \mathcal{C}_j=\emptyset$ for $i\neq j$. During task $t$, the model can only access $\mathcal{D}_t$ and cannot replay samples from previous tasks. After learning task $t$, the model must classify test samples from all seen classes $\mathcal{Y}_t=\bigcup_{i=1}^{t}\mathcal{C}_i$.

We build CLARE as a parameter-efficient CIL method on top of a pretrained vision transformer (ViT-B/16 \citep{dosovitskiy2020vit}). The pretrained backbone parameters are denoted by $\theta_0$ and kept frozen. We insert lightweight adapters \citep{chen2022adaptformer} into each transformer block and denote all adapter parameters by $\alpha$. The classifier after task $t$ is denoted by $\phi_t$, and the prediction function is written as $f(x;\theta_0,\alpha,\phi_t)$. CLARE only updates the adapters and classifier. It does not fine-tune the full backbone.

\begin{figure*}[t]
    \centering
    \includegraphics[width=0.9\textwidth]{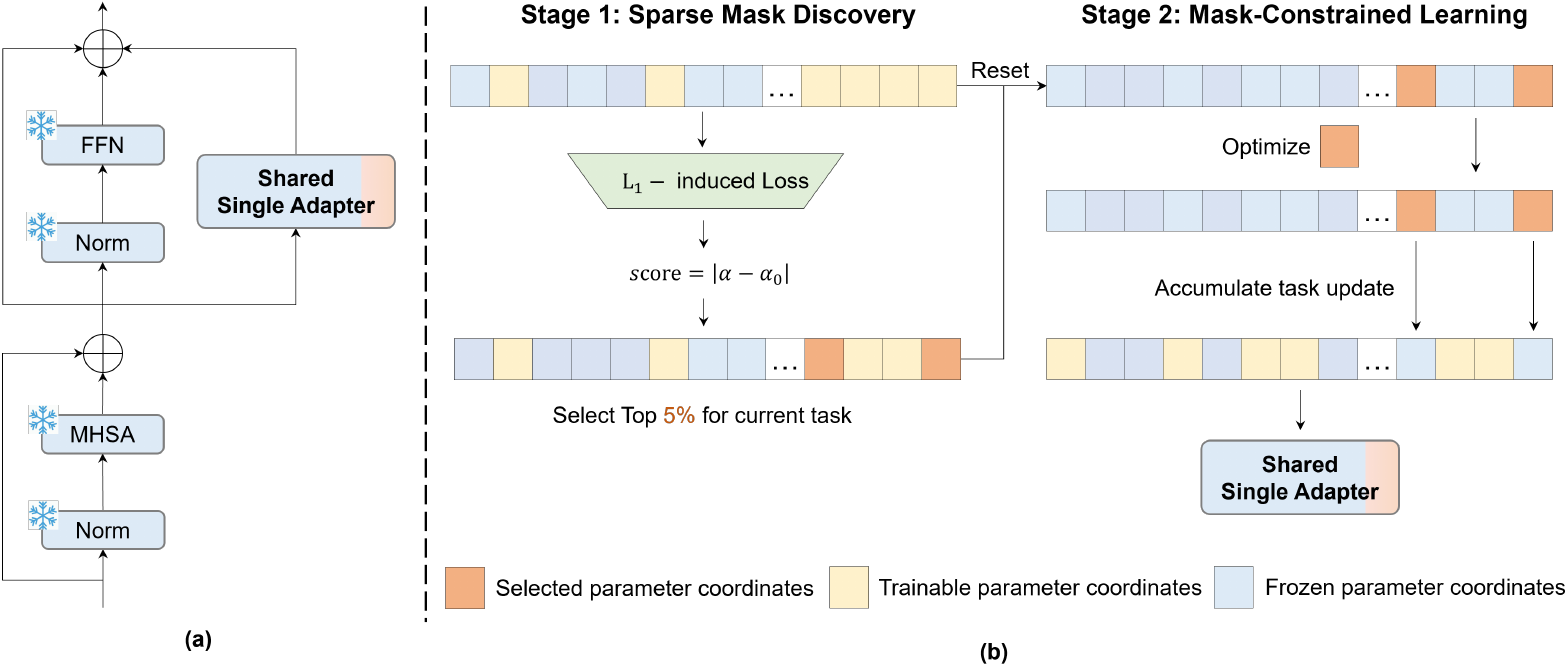}
    \vspace{1em}
    \caption{\textbf{(a)} A shared adapter contains parameters that have been frozen after previous tasks (blue) and parameters that remain available for future tasks (orange). \textbf{(b)} For a new task, CLARE first selects a sparse set of parameters from the available pool (Stage 1), then resets them to their original values and optimizes only those selected parameters (Stage 2). After training, the updated parameters are frozen and added to the frozen set.}
    \label{fig:method}
\end{figure*}

Figure~\ref{fig:method} gives an overview. CLARE keeps one shared adapter across the whole task sequence. For each task, it first explicitly learns a parameter-importance sparse mask for the current task using an $L_1$-regularized objective. Subsequently, it trains only the selected adapter parameters. After task learning, the masked task update is added to the shared adapter, and the next task starts from this updated adapter.

\subsection{Capacity-Aware Sparse Adapter Learning}
CLARE uses one shared adapter to learn all tasks. This design is parameter-efficient, but it may create a clear source of forgetting. 
Specifically, if a new task changes adapter parameters that are important for old tasks, the feature representation of historical classes can shift. The classifier was trained on the historical representations, so this shift can damage old-task predictions and cause catastrophic forgetting. 
The problem becomes more severe in long task sequences because more tasks compete for the same limited trainable parameters.

To control this interference, CLARE explicitly manages which scalar entries of the adapter parameters can be updated by each task. We call each scalar entry an \textit{adapter coordinate}. Let $\Omega=\{1,\ldots,N\}$ index all adapter coordinates, where $N$ is the total number of adapter parameters. After several tasks have been learned, some coordinates have already been selected by previous task masks. For task $t$, let $\Omega_{<t}=\bigcup_{i=1}^{t-1}\Omega_i$ be the set of coordinates used by previous tasks, and let
\begin{equation}
\label{eq:a_t}
\mathcal{A}_t = \Omega \setminus \Omega_{<t}
\end{equation}
be the set of coordinates still available for task $t$. CLARE selects a sparse set $\Omega_t\subseteq\mathcal{A}_t$ for the current task. In this way, new tasks are guided toward unused adapter parameters instead of freely overwriting parameters already assigned to earlier tasks.

Importantly, since knowledge of all historical tasks are learned in $\Omega_{<t}$, we encourage $\Omega_{t}$ to build on the accumulated adapter state from previous tasks and use minimal sparse updates for the new task $t$. Specifically, the percentage of $\mathcal{A}_t$ that can be updated for task $t$ (sparsity ratio) is set to a fixed value.
For example, $\rho=0.95$ means that task $t$ only uses $1-\rho=5\%$ of $\mathcal{A}_t$ to learn, not $5\%$ of the original adapter capacity for every task. 
If $F_t=|\mathcal{A}_t|$ is the number of available coordinates before task $t$, then
\begin{equation}
F_{t+1}\approx \rho F_t, \qquad F_t\approx \rho^{t-1}N.
\end{equation}
After $T$ tasks, the used capacity is approximately $1-\rho^T$. For $\rho=0.95$ and $T=100$, this value is $1-0.95^{100}\approx 0.994$. The adapter is nearly saturated after 100 tasks, but the method does not suffer from interference where each parameter coordinate is used by $5\%\times100=5$ different tasks on average when assigning a fixed 5\% budget to every task. 
This geometric schedule slows capacity consumption and explains why a single adapter can support a 100-task sequence. 

Let $\alpha_0$ denote the initial adapter parameters before any continual task is learned. For task $t$, CLARE learns an adapter state $\alpha_t$ and defines the task-specific adapter update as
\begin{equation}
\Delta\alpha_t = \alpha_t - \alpha_0.
\end{equation}
Sparsity is imposed on $\Delta\alpha_t$, not on the full pretrained backbone. 
Since a task modifies only free coordinates $\mathcal{A}_t$ in Equation~\ref{eq:a_t} while used ones are frozen. In this trainable subspace, the previous shared adapter equals $\alpha_0$, so defining $\Delta\alpha_t$ w.r.t.\ $\alpha_0$ or the previous shared state is equivalent.

After each task, the sparse update is immediately added to the shared adapter. 
For tasks after the first one, CLARE starts from this learned shared adapter instead of returning to an independent adapter. 
The next task can therefore use the representation accumulated so far, while the free-coordinate mask limits direct overwriting of earlier task updates.

\subsection{Two-Stage Sparse Update Optimization}
We now describe how the sparse adapter coordinate set $\Omega_t$ and the adapter update $\Delta\alpha_t$ are learned for each task $t$. CLARE uses a two-stage approach: the first stage selects $\Omega_t$ from the currently available coordinates, and the second stage optimizes $\Delta\alpha_t$ while restricting adapter updates to $\Omega_t$.

\textbf{Stage 1: $L_1$-induced mask discovery.}
For task $t$, the first stage decides which available adapter coordinates should be assigned to the current task. CLARE starts from the current shared adapter and updates only the coordinates in $\mathcal{A}_t$ for a few epochs using the current task data. The objective is
\begin{equation}
\min_{\alpha}\;
\mathbb{E}_{(x,y)\sim\mathcal{D}_t} \left[\ell\left(f(x;\theta_0,\alpha,\phi_t),y\right)\right]
+
\lambda
\left\|P_{\mathcal{A}_t}(\alpha-\alpha_0)\right\|_1,
\label{eq:mask_loss}
\end{equation}
where $P_{\mathcal{A}_t}(\cdot)$ keeps only currently free adapter coordinates and $\lambda$ controls the strength of the sparsity term. The classifier provides the task loss, while the purpose of this stage is to reveal which adapter coordinates respond to the new task. The $L_1$ penalty encourages most free-coordinate updates to stay close to zero. As a result, a coordinate that still changes by a large amount is likely to be important for learning task $t$, while coordinates with near-zero changes can be left unused.

Using the adapter values reached at the end of this short optimization, CLARE scores each available coordinate by the magnitude of this movement:
\begin{equation}
s_t^{(j)} = \left|\alpha^{(j)}-\alpha_0^{(j)}\right|, \qquad j\in\mathcal{A}_t.
\end{equation}
Larger scores mean that the coordinate moved more even under the $L_1$ penalty, so CLARE treats them as more useful for the current task.

The number of coordinates selected for task $t$ follows the remaining-capacity schedule in the previous subsection. The target budget is
\begin{equation}
\bar{k}_t=\mathrm{round}\left((1-\rho)|\mathcal{A}_t|\right).
\end{equation}
Since $\bar{k}_t$ may be zero after rounding when few coordinates remain, CLARE clips it to a valid integer budget:
\begin{equation}
k_t=\min\left\{|\mathcal{A}_t|,\max\left\{1,\bar{k}_t\right\}\right\},
\qquad \text{when } |\mathcal{A}_t|>0.
\label{eq:num_selected_coordinates}
\end{equation}
If no available coordinate remains, CLARE assigns an empty mask. Otherwise, CLARE ranks all scores $\{s_t^{(j)}:j\in\mathcal{A}_t\}$ globally and sets $\tau_t$ to the $k_t$-th largest score. The selected coordinate set is
\begin{equation}
\Omega_t = \left\{j\in\mathcal{A}_t: s_t^{(j)}\ge\tau_t\right\}.
\label{eq:coordinate_selection}
\end{equation}
Equivalently, the binary mask is $M_t^{(j)}=\mathbb{I}[j\in\Omega_t]$. If several coordinates have exactly the same score as $\tau_t$, they are all included, so the mask size can differ slightly from $k_t$ in the rare case of ties. This global top-$k_t$ rule has two purposes. First, it gives each task only a fixed fraction of the coordinates still available, which preserves capacity for future tasks. Second, it lets the task use that capacity wherever the learned update indicates it is most useful, instead of forcing every adapter layer to receive the same budget. Excluding $\Omega_{<t}$ further prevents direct reuse of coordinates already assigned to previous tasks.

\textbf{Stage 2: mask-constrained learning.}
The adapter values obtained in Stage 1 are not used as the final task parameters. CLARE restores the adapter to the state used at the beginning of Stage 1: the initial adapter for the first task and the shared adapter learned so far for later tasks. It then learns task $t$ while updating only the coordinates in $\Omega_t$. Equivalently, for any adapter coordinate $j$ not selected in Stage 1, CLARE sets the gradient of $\alpha_t^{(j)}$ to zero before each optimizer step:
\begin{equation}
\begin{aligned}
&\min_{\alpha_t,\phi_t}\;
\mathbb{E}_{(x,y)\sim\mathcal{D}_t}
\left[\ell_{\mathrm{cls}}\left(f(x;\theta_0,\alpha_t,\phi_t),y\right)\right],\\
&\nabla_{\alpha_t^{(j)}}=0 \quad \text{if } j\notin\Omega_t.
\end{aligned}
\label{eq:masked_training}
\end{equation}
Here $\nabla_{\alpha_t^{(j)}}$ denotes the gradient with respect to the $j$-th scalar adapter coordinate after the task loss is back-propagated. Only coordinates in $\Omega_t$ receive adapter updates, while all other adapter coordinates remain fixed. This reset is important: Stage 1 is designed to select coordinates under an $L_1$ pressure, whereas Stage 2 is designed to learn the task well once the coordinate budget has been fixed. Removing the $L_1$ term in Stage 2 prevents the sparsity objective from weakening classification learning. In practice, the classifier is expanded for the new classes and trained together with the active adapter coordinates.

\subsection{Sequential Sparse Update and Inference}
After task learning, CLARE stores the masked adapter delta
\begin{equation}
\Delta\alpha_t^{\mathrm{sparse}}
=
M_t\odot(\alpha_t-\alpha_0).
\label{eq:sparse_delta}
\end{equation}
where $M_t^{(j)}=\mathbb{I}[j\in\Omega_t]$. Thus, only the coordinates selected for task $t$ are written into the task update. All other coordinates contribute zero.
The shared adapter used after task $t$ is obtained by adding the sparse task deltas to the initial adapter state:
\begin{equation}
\alpha_{\mathrm{shared}}^{(t)}
=
\alpha_0+
\sum_{i=1}^{t}\Delta\alpha_i^{\mathrm{sparse}}.
\label{eq:adapter_merge}
\end{equation}
Because each task selects coordinates from the remaining free capacity, the shared adapter accumulates task knowledge while reducing direct overlap among task updates. The update is performed sequentially after each task, so the next task always starts from the adapter learned so far.

At inference time, CLARE uses the frozen backbone $\theta_0$, the shared adapter $\alpha_{\mathrm{shared}}^{(t)}$, and a single classifier $\phi_t$ over all seen classes:
\begin{equation}
\hat{y}=
\arg\max_{c\in\mathcal{Y}_t}
f_c(x;\theta_0,\alpha_{\mathrm{shared}}^{(t)},\phi_t).
\end{equation}
Throughout continual learning, we maintain only one set of adapter weights and do not require task identity for inference.
We use a cosine classifier whose class weights grow as new classes arrive following standard implementations~\citep{zhou2024EASE}. 

CLARE differs from existing sparse and parameter-efficient continual learning methods in two ways: (1) Its sparsity is induced during learning through a $L_1$-regularization.  
(2) The mask is capacity-aware since each task selects from the remaining free coordinates. 
These properties allow CLARE to use a single adapter to scale to long task sequences.
\vspace{-0.5em}

\section{Experiments}
\label{sec:experiments}

We evaluate CLARE in exemplar-free class-incremental learning (CIL) under two settings ranging from 10 tasks to 100 tasks. The main setting studies the challenging long task sequence setup, where methods must preserve performance as the number of incremental tasks grows to 100. The second setting leverages standard CIL protocols on established datasets to verify that the same sparse shared-adapter design in CLARE remains competitive under shorter sequences. We then comprehensively investigate the effect of each component of CLARE through controlled ablation studies.

\subsection{Experimental Setup}
\label{sec:experimental_setup}

\textbf{Datasets.} Long-sequence evaluation uses ImageNet-R \citep{hendrycks2021ImagenetR} and ImageNet-A \citep{hendrycks2021ImagenetA} split into 50 tasks with 4 classes per task, and OmniBenchmark-1k \citep{lou2026CARE} split into 100 tasks with 10 classes per task. We denote Inc$N$ as the number of novel classes learned per task. Following previous works \citep{zhou2024EASE,zhou2025APER,liang2024inflora,Wu2025SDLoRA}, standard benchmark evaluation uses ImageNet-R \citep{hendrycks2021ImagenetR} with 10 tasks (Inc20) and 20 tasks (Inc10), CIFAR-100 \citep{krizhevsky2009CIFAR100} with 10 tasks (Inc10) and 20 tasks (Inc5), ImageNet-A \citep{hendrycks2021ImagenetA} with 10 tasks (Inc20), and ObjectNet with 10 tasks (Inc20).

\noindent\textbf{Baselines.} We compare with representative pretrained-model CIL methods from different families, including prompt-based (L2P \citep{wang2022L2P}, DualPrompt \citep{wang2022dualprompt}, CODA-Prompt \citep{smith2023coda}), adapter-based \citep{zhou2024EASE, wang2025sema,gao2025MoAL,zhou2025APER}, classifier-based \citep{zhang2023slca,goswami2023fecam} and LoRA-based \citep{liang2024inflora,Wu2025SDLoRA}.
These baselines cover methods that use task-specific parameters or prompts, and methods that maintain a single model at inference time. All adapter-based and LoRA-based baselines are run from their official implementations.

\noindent\textbf{Implementation details.} CLARE follows prior works~\citep{zhou2024EASE,zhou2025APER,liang2024inflora,Wu2025SDLoRA} to use a ViT-B/16 pretrained on ImageNet-21K as the frozen backbone and trains lightweight adapters together with the classifier. 
During training, we use a learning rate of 0.02 with cosine learning rate decay, a batch size of 32, and a SGD optimizer with weight decay of 0.0005.
The adapter bottleneck dimension is 64, the first-stage mask discovery is trained for 5 epochs with an $L_1$ coefficient of $10^{-4}$, and the second-stage masked training is run for 20 epochs. The sparsity ratio is set to 95\%.

\noindent\textbf{Metrics.} We report the average incremental accuracy $\bar{A}$ and the final accuracy $A_T$. Here, $A_T$ is the accuracy over all seen classes after training the final task, and $\bar{A}$ is the mean of the accuracies measured after each incremental task. Higher values are better for both metrics.

\subsection{Long Task Sequence Evaluation}
\label{exp:long_sequence}

\begin{table*}[t]
  \centering
  \caption{Performance comparison on benchmarks with long task sequences. We report average incremental accuracy $\bar{A}$ and final accuracy $A_T$.}
  \vspace{1em}
  \scalebox{0.82}{
    \begin{tabular}{l|cc|cc|cc}
    \toprule
    \multirow{3}[2]{*}{Method}
      & \multicolumn{2}{c|}{\textbf{ImageNet-R}}
      & \multicolumn{2}{c|}{\textbf{ImageNet-A}}
      & \multicolumn{2}{c}{\textbf{OmniBenchmark-1k}} \\
      & \multicolumn{2}{c|}{\textbf{50 Tasks (Inc4)}}
      & \multicolumn{2}{c|}{\textbf{50 Tasks (Inc4)}}
      & \multicolumn{2}{c}{\textbf{100 Tasks (Inc10)}} \\
      & $\mathbf{\bar{A}}$ & $\mathbf{A}_T$
      & $\mathbf{\bar{A}}$ & $\mathbf{A}_T$
      & $\mathbf{\bar{A}}$ & $\mathbf{A}_T$\\
    \midrule
    L2P\textsubscript{(CVPR'22)}          & 69.16 & 63.45 & 49.89 & 36.41 & 60.91 & 48.87 \\
    DualPrompt\textsubscript{(ECCV'22)}   & 64.00 & 56.33 & 43.85 & 29.95 & 62.18 & 49.45 \\
    CODA-Prompt\textsubscript{(CVPR'23)}  & 62.43 & 57.57 & 38.24 & 26.60 & 64.16 & 51.75 \\
    EASE\textsubscript{(CVPR'24)}         & 78.11 & 70.63 & 59.86 & 47.53 & 65.00 & 53.54 \\
    SEMA\textsubscript{(CVPR'25)}         & 67.80 & 59.32 & 52.99 & 40.68 & 56.55 & 33.96 \\
    APER-Adapter\textsubscript{(IJCV'25)} & 72.43 & 64.83 & 61.29 & 48.58 & 73.23 & 62.24 \\
    InfLoRA\textsubscript{(CVPR'24)}      & 71.68 & 62.62 & 50.17 & 38.70 & 51.53 & 27.01 \\
    SD-LoRA\textsubscript{(ICLR'25)}      & 68.40 & 63.28 & 54.72 & 41.28 & 53.97 & 28.15 \\
    \rowcolor[rgb]{0.8, 0.9, 0.95}\textbf{CLARE (Ours)} & \textbf{83.00} & \textbf{76.52} & \textbf{68.83} & \textbf{57.98} & \textbf{78.32} & \textbf{66.88} \\
    \bottomrule
    \end{tabular}%
  }
  \label{tab:long_eval_2}%
  \vspace{-0.75em}
\end{table*}

Table~\ref{tab:long_eval_2} evaluates long task sequences, a setting where single adapter-based models that do not leverage task identity at inference struggle (e.g., InfLoRA, SD-LoRA). CLARE consistently achieves the highest final accuracy and average incremental accuracy. On the longest OmniBenchmark-1k benchmark comprising 100 tasks, CLARE attains a final accuracy of 66.88\%, which represents a substantial 137\% relative improvement over SD-LoRA \citep{Wu2025SDLoRA}. This gain suggests that the capacity-aware sparsity introduced by CLARE can mitigate catastrophic forgetting even for 100 tasks. 
Compared to Aper \citep{zhou2025APER}, which trains the adapter only on the first task, freezes it for subsequent tasks, and relies on class prototypes for prediction, CLARE yields significant improvements of 11.69\% on 50-task ImageNet-R and 9.4\% on 50-task ImageNet-A, respectively.
This demonstrates stronger learning plasticity of CLARE.
These results indicate that CLARE can maintain a balance between learning plasticity and knowledge retention even over long task sequences. 

\subsection{Standard Benchmark Evaluation}
\label{sec:standard_benchmark_eval}

\begin{table*}[t]
  \centering
  \caption{Performance comparison with state-of-the-art methods on standard CIL benchmarks. We report average incremental accuracy $\bar{A}$ and final accuracy $A_T$.}
  \vspace{1em}
  \resizebox{0.985\textwidth}{!}{
    \begin{tabular}{l|cccc|cccc|cc|cc}
    \toprule
    \multirow{3}[2]{*}{Method} & \multicolumn{4}{c|}{\textbf{ImageNet-R}} & \multicolumn{4}{c|}{\textbf{CIFAR-100}} & \multicolumn{2}{c|}{\textbf{ImageNet-A}} & \multicolumn{2}{c}{\textbf{ObjectNet}} \\
          & \multicolumn{2}{c}{\textbf{10 Tasks (Inc20)}} & \multicolumn{2}{c|}{\textbf{20 Tasks (Inc10)}} & \multicolumn{2}{c}{\textbf{10 Tasks (Inc10)}} & \multicolumn{2}{c|}{\textbf{20 Tasks (Inc5)}} & \multicolumn{2}{c|}{\textbf{10 Tasks (Inc20)}} & \multicolumn{2}{c}{\textbf{10 Tasks (Inc20)}}  \\
          & $\mathbf{\bar{A}}$ & $\mathbf{{A}}_T$ & $\mathbf{\bar{A}}$ & $\mathbf{{A}}_T$ & $\mathbf{\bar{A}}$ & $\mathbf{{A}}_T$ & $\mathbf{\bar{A}}$ & $\mathbf{{A}}_T$ & $\mathbf{\bar{A}}$ & $\mathbf{{A}}_T$ & $\mathbf{\bar{A}}$ & $\mathbf{{A}}_T$  \\
    \midrule
    L2P\textsubscript{(CVPR'22)} & 75.46 & 69.77 & 63.75 & 55.78 & 85.92 & 79.19 & 85.94 & 79.93 & 49.39 & 41.71 & 66.77 & 55.16 \\
    DualPrompt\textsubscript{(ECCV'22)} & 73.10 & 67.18 & 66.52 & 61.77 & 89.65 & 84.89 & 87.87 & 81.15 & 53.71 & 41.67 & 64.31 & 52.99 \\
    CODA-Prompt\textsubscript{(CVPR'23)} & 77.97 & 72.27 & 70.45 & 64.68 & 91.05 & 86.44 & 89.11 & 81.96 & 53.54 & 42.73 & 66.53 & 56.80 \\
    EASE\textsubscript{(CVPR'24)} & 81.74 & 76.17 & 81.18 & 74.62 & 92.11 & 87.72 & 91.51 & 85.80 & 65.34 & 55.04 & 71.04 & 59.37 \\
    SEMA\textsubscript{(CVPR'25)} & 81.39 & 77.84 & 77.84 & 69.60 & 91.60 & 86.75 & 92.23 & 87.84 & 63.83 & 52.21 & 67.95 & 54.92 \\
    APER-Adapter\textsubscript{(IJCV'25)} & 75.82 & 67.95 & 72.35 & 64.33 & 92.22 & 87.45 & 90.65 & 85.15 & 60.53 & 49.57 & 69.24 & 57.41 \\
    InfLoRA\textsubscript{(CVPR'24)} & 80.82 & 75.65 & 77.28 & 71.01 & 91.70 & 86.51 & 89.13 & 81.46 & 58.50 & 46.28 & 70.67 & 58.04 \\
    SD-LoRA\textsubscript{(ICLR'25)} & 82.04 & 77.34 & 80.22 & 75.26 & 92.54 & 88.01 & 90.90 & 85.18 & 64.95 & 55.96 & 70.37 & 58.54 \\
    \rowcolor[rgb]{0.8, 0.9, 0.95}\textbf{CLARE (Ours)} & \textbf{83.88} & \textbf{79.73} & \textbf{82.67} & \textbf{77.73} & \textbf{94.73} & \textbf{91.92} & \textbf{94.96} & \textbf{91.73} & \textbf{69.46} & \textbf{61.36} & \textbf{76.67} & \textbf{67.77} \\
    \bottomrule
    \end{tabular}
    }
  \label{tab:standard_benchmark}%
  \vspace{-0.75em}
\end{table*}

Table~\ref{tab:standard_benchmark} evaluates CLARE under the commonly used CIL settings with 10 and 20 tasks~\citep{wang2022L2P,smith2023coda,zhou2024EASE,liang2024inflora,Wu2025SDLoRA,wang2025sema}. 
Across all six reported settings over four datasets, CLARE achieves the highest final accuracy among the compared methods. 
The gains are especially clear on the datasets with larger distribution shifts: on ImageNet-A Inc20, CLARE improves $A_T$ from 55.96\% for the strongest baseline (SD-LoRA) to 61.36\%, and on ObjectNet Inc20, it improves $A_T$ from 59.37 to 67.77. The results on ImageNet-R and CIFAR-100 show a similar pattern under both 10 and 20 task splits. 
On ImageNet-R, CLARE improves final accuracy over the strongest baseline by 2.47\% in the 20-task Inc10 setting. On CIFAR-100 with 10 tasks, the improvement is 3.91\%.
Together with Table~\ref{tab:long_eval_2}, these consistent gains across different number of tasks indicate that constraining adapter updates to sparse, selected coordinates in CLARE is a scalable and can generalize to various ranges of tasks.

\subsection{Ablation Studies}
\label{sec:ablation}

\begin{table*}[t]
  \centering
  \caption{Impact of capacity-aware sparsity and two stage learning on ImageNet-R with 10 tasks and Omnibenchmark-1K with 100 tasks.}
  \vspace{1em}
  \setlength{\tabcolsep}{8pt}
  \scalebox{0.82}{
    \begin{tabular}{lccc|ccc}
    \toprule
    & \multicolumn{3}{c|}{\textbf{ImageNet-R}} & \multicolumn{3}{c}{\textbf{Omnibenchmark-1K}} \\
    Variant & $\bar{A}$ & $A_T$ & $\Delta A_T$ & $\bar{A}$ & $A_T$ & $\Delta A_T$ \\
    \midrule
    \rowcolor[rgb]{0.8, 0.9, 0.95} CLARE & 83.88 & 79.73 & 0.00 & 78.32 & 66.88 & 0.00\\
    w/o capacity-aware sparsity & 83.94 & 79.60 & -0.13 & 64.20 & 53.17 & -13.71 \\
    w/o two-stage learning & 83.25 & 78.96 & -1.08 & 75.19 & 63.09 & -3.79 \\
    \bottomrule
    \end{tabular}
  }
  \label{tab:ablation_components}
  \vspace{-0.75em}
\end{table*}

We investigate each design component of CLARE, which are the capacity-aware sparse adapter learning for determining how many parameters to use for the current task and two-stage sparse update optimization for maximizing learning plasticity using the amount of parameters available.
Table~\ref{tab:ablation_components} shows results on ImageNet-R with 10 tasks and Omnibenchmark-1k with 100 tasks. In the second row of Table~\ref{tab:ablation_components}, we enforce a constant predefined portion of parameters for each task. We report the best results from over four different sparsity ratios \{10\%, 5\%, 2.5\%, 1\%\} for both datasets.
It can be observed that on the 100 task setting, the final accuracy drops significantly by 13.71\% to 53.17\%, demonstrating the importance of capacity-aware sparsity.
In the third row (w/o two-stage), we remove the two-stage learning process and directly train the adapter, and select the parameters for each task based on its magnitude. It can be seen that final accuracy decreased for both settings, suggesting that the two stage optimization is beneficial.

We further investigate design choices and important hyperparameters within each component. Specifically, we study the effect of the sparsity ratio, the training strategies for the two-stage sparse update optimization, and the effect of the $L_1$ normalization coefficient $\lambda$.

\paragraph{Impact of the Sparsity Ratio}.
Table~\ref{tab:ablation_sparsity_ratio} shows the impact of the sparsity ratio for 10 tasks (ImageNet-R) and 100 tasks (Omnibenchmark-1k). We observe that CLARE is robust to the sparsity ratio when choosing from 95\% to 97.5\%, where the declines in final accuracy are not substantial (-0.38\% for 10 tasks and -0.41\% for 100 tasks). We note that choosing the sparsity ratio to be smaller (90\%) can improve performance on the short 10 task setting, but lead to substantial decline in performance in the 100 task setting. This decline is due to insufficient number of parameters to learn new task when the number of tasks increases to 100.

\begin{table}[t]
  \centering
  \caption{Impact of the sparsity ratio.}
  \vspace{1em}
  \setlength{\tabcolsep}{8pt}
  \scalebox{0.82}{
    \begin{tabular}{lccc|ccc}
    \toprule
    & \multicolumn{3}{c|}{\textbf{ImageNet-R}} & \multicolumn{3}{c}{\textbf{Omnibenchmark-1K}} \\
    Variant & $\bar{A}$ & $A_T$ & $\Delta A_T$ & $\bar{A}$ & $A_T$ & $\Delta A_T$ \\
    \midrule
    90\% & 83.88 & 84.04 & +0.45 & 73.49 & 59.25 & -7.63 \\
    \rowcolor[rgb]{0.8, 0.9, 0.95} 95\% & 83.88 & 79.73 & 0.00 & 78.32 & 66.88 & 0.00\\
    97.5\% & 82.38 & 79.35 & -0.38 & 79.03 & 66.47 & -0.41 \\
    99\% & 79.62 & 73.14 & -6.59 & 77.62 & 65.59 & -1.29 \\
    \bottomrule
    \end{tabular}
  }
  \label{tab:ablation_sparsity_ratio}
  \vspace{-0.75em}
\end{table}

\begin{table*}[t]
  \centering
  \caption{Impact of mask selection strategies, two-stage training, and $L_1$ coefficient $\lambda$ on ImageNet-R with 10 tasks and Omnibenchmark-1K with 100 tasks. independent tuning trains each task from the initial (not shared) adapter, then merges.}
  \vspace{1em}
  \setlength{\tabcolsep}{8pt}
  \scalebox{0.82}{
    \begin{tabular}{lccc|ccc}
    \toprule
    & \multicolumn{3}{c|}{\textbf{ImageNet-R}} & \multicolumn{3}{c}{\textbf{Omnibenchmark-1K}} \\
    Variant & $\bar{A}$ & $A_T$ & $\Delta A_T$ & $\bar{A}$ & $A_T$ & $\Delta A_T$ \\
    \midrule
    \rowcolor[rgb]{0.8, 0.9, 0.95} CLARE ($\lambda=0.0001$) & 83.88 & 79.73 & 0.00 & 78.32 & 66.88 & 0.00\\
    \midrule
    No sparse mask & 24.36 & 0.23 & -79.50 & 8.73 & 0.04 & -66.84 \\
    Random mask & 83.16 & 79.22 & -0.51 & 70.10 & 61.23 & -5.65  \\
    Independent tuning & 58.32 & 32.33 & -47.40 & 23.95 & 11.03 & -55.85 \\
    \midrule 
    Stage-1 L2 regularization & 83.96 & 79.59 & -0.20 & 76.17 & 65.36 & -1.52 \\
    Stage-1 $\lambda=0$ & 83.05 & 78.85 & -0.88 & 75.84 & 64.97  & -1.91 \\
    Stage-1 $\lambda=0.00001$ & 84.08 & 79.92 & +0.19 & 77.34 & 65.69 & -1.19 \\
    Stage-1 $\lambda=0.001$ & 84.29 & 79.61 & -0.12 & 76.92 & 66.16 & -0.58 \\
    \bottomrule
    \end{tabular}
  }
  \label{tab:ablation_two_stage}
  \vspace{-0.75em}
\end{table*}

\paragraph{Training Strategies for Two-Stage Learning}.
Table~\ref{tab:ablation_two_stage} examines the effect of different mask selection and optimization strategies, as well as the sensitivity to the sparsity coefficient $\lambda$. Removing the sparse mask entirely (no sparse mask) causes severe forgetting, with final accuracy collapsing to near zero on both benchmarks, confirming that restricting adapter updates is essential for retaining previous knowledge. Replacing the learned mask with a random selection of the same number of available coordinates (random mask) reduces final accuracy by 0.51\% on ImageNet-R and 5.65\% on OmniBenchmark-1k, indicating that the $L_1$-guided mask identifies task-relevant coordinates are beneficial. 
When each task is trained independently from the initial pretrained adapter, without leveraging prior accumulated knowledge in the shared adapter (independent tuning), performance drops drastically to 32.33\% and 11.03\% final accuracy, respectively. 
This collapse shows that knowledge accumulation across tasks is critical and that sparse masking alone is insufficient without a shared representation.

We further study alternatives for the first stage of the two-stage optimization. Using $L_2$ regularization instead of $L_1$ (Stage-1 $L_2$) yields a modest decline, notably a 1.52\% lower final accuracy on the 100-task setting, suggesting that the sparsity-inducing property of $L_1$ is better suited for selecting a compact set of coordinates.
Removing the sparsity term entirely ($\lambda=0$) and selecting coordinates solely by update magnitude after unregularized probing reduces final accuracy by 0.88\% and 1.91\%, respectively, confirming that the $L_1$ penalty helps identify the most essential adapter parameters for each task. 
Varying $\lambda$ around the default value of $0.0001$ ($\lambda=0.00001$ and $\lambda=0.001$) yields stable final accuracies, demonstrating that the method is robust to the choice of this hyperparameter.

\paragraph{Robustness to Seeds}

In the main experiments and ablations of the paper, we partition each dataset by randomly shuffling the class order using the seed 1993, following EASE. In Table~\ref{tab:seed}, we evaluate the performance of CLARE and a representative baseline SD-LoRA on three random seed and report the mean$\pm$std. We observe that variance is small and final accuracy is close to that of the seed 1993 so CLARE is robust to different orders across different benchmarks.

\begin{table*}[t]
\centering
\caption{Final accuracy $A_T$ (mean$\pm$std over 3 random seeds).}
\vspace{1em}
  \setlength{\tabcolsep}{8pt}
  \scalebox{0.82}{
\begin{tabular}{lccc}
\toprule
Method & ImageNet-R 10 Task & ImageNet-A 50 Task & Omnibenchmark-1k 100 Task \\
\midrule
\rowcolor[rgb]{0.8, 0.9, 0.95} CLARE & 79.48{\scriptsize$\pm$0.45} & 57.49{\scriptsize$\pm$0.62} & 67.10{\scriptsize$\pm$0.46} \\
SD-LoRA & 77.50{\scriptsize$\pm$0.38} & 41.33{\scriptsize$\pm$1.91} & 13.27{\scriptsize$\pm$12.83} \\
\bottomrule
\end{tabular}}
\label{tab:seed}
\vspace{-2mm}
\end{table*}

\paragraph{Impact of Adapter Capacity}.
The adapter bottleneck dimension $d$ controls the total number of trainable adapter coordinates, i.e., the size of the pool from which each task selects its sparse mask. We investigate the impact of $d$ on CLARE's gains on long task sequences. Specifically, we vary $d\in\{16,32,64,128\}$ on ImageNet-R with 50 tasks.

\begin{table}[t]
  \centering
  \caption{Impact of adapter bottleneck dimension $d$ on ImageNet-R with 50 tasks (Inc4). Relative width is measured with respect to the default $d=64$.}
  \vspace{1em}
  \setlength{\tabcolsep}{8pt}
  \scalebox{0.82}{
    \begin{tabular}{lccc}
    \toprule
    Variant & Relative width & $A_T$ & $\Delta A_T$ \\
    \midrule
    $d=16$ & $\times 0.25$ & 76.41 & -0.11 \\
    $d=32$ & $\times 0.50$ & 76.38 & -0.14 \\
    \rowcolor[rgb]{0.8, 0.9, 0.95} $d=64$ & $\times 1.00$ & 76.52 & 0.00 \\
    $d=128$ & $\times 2.00$ & 76.65 & +0.13 \\
    \bottomrule
    \end{tabular}
  }
  \label{tab:ablation_adapter_capacity}
  \vspace{-0.75em}
\end{table}

Table~\ref{tab:ablation_adapter_capacity} shows that final accuracy is nearly insensitive to adapter width. Reducing the bottleneck to $d=16$ (one quarter of the default width) changes $A_T$ by only $-0.11$ points, while doubling it to $d=128$ improves $A_T$ by only $+0.13$ points. Across this eight-fold range, the spread in final accuracy is $0.27$ points, and even $d=16$ remains well above the strongest baselines on the same split in Table~\ref{tab:long_eval_2} (EASE $70.63$, InfLoRA $62.62$).
This 50-task setting is a stringent test of capacity: with $\rho=0.95$, about $1-0.95^{50}\approx 92\%$ of the adapter has already been assigned, so extra width would help if the method were limited by the number of parameters. The nearly unchanged $A_T$ indicates that it is not. CLARE's improvements instead come from assigning each task a sparse subset of the remaining coordinates, which reduces destructive interference even when the adapter itself is small.

\subsection{Efficiency} CLARE keeps a single shared adapter and does not use task identity or routing at inference, so it adds no inference-time overhead relative to a standard adapter model. Although Stage~1 spends a few epochs on mask discovery, Stage~2 updates only the selected sparse coordinates, which keeps training efficient. On 50-task ImageNet-A, the total training time of CLARE is $4.37$ hours on a single NVIDIA L40 GPU, slightly faster than SD-LoRA ($4.41$ hours) under the same setting.

\section{Limitations}
\label{sec:limitations}
CLARE is designed so that the sequence length $T$ need not be known in advance. Rather than splitting the adapter into a predefined number of task slots, each new task selects a sparse subset of the coordinates that have not yet been assigned. The consumed capacity therefore grows as $1-\rho^{T}$. With the default $\rho=0.95$, a single adapter is only about $99.4\%$ full after $100$ tasks, which is enough to remain accurate from the standard $10$--$20$ task setting through the $100$-task OmniBenchmark-1k protocol. The limitation is that this pool is finite and cannot scale indefinitely to very large number of tasks like 10k tasks. On substantially longer streams the remaining free coordinates will run out, and new tasks would then have little unused capacity left to learn. 
Expanding capacity at that point does not require changing the allocation principle. One can freeze a saturated adapter and continue with a fresh one, or increase the bottleneck when free coordinates become scarce, then apply the same remaining-capacity schedule. We leave such expansion to future work.
Notably, prior works like SD-LoRA mostly focus on 10 to 20 tasks, while our proposed CLARE can learn effectively for 100 tasks.


\section{Conclusion}
In this paper, we have addressed the challenge of scaling class-incremental continual learning to long task sequences by proposing CLARE, a sparsity-driven adapter-based framework.
CLARE is built on a simple yet effective principle, where each task updates only a sparse subset of parameters, selected from parameters not already assigned to earlier tasks. 
The sparse updates are accumulated as it learns new tasks.
This design eliminates the need for task-specific routing or stored task-specific parameters at inference, making it particularly suitable for long task sequences where storing per-task modules becomes impractical.
Extensive experiments on both long-sequence and standard benchmarks demonstrate that CLARE consistently outperforms representative continual learning methods.

Several limitations point to directions for future work.
For instance, the geometric capacity schedule inevitably saturates the adapter as the number of tasks increases, limiting the maximum sequence length that the current method can support without expanding the adapter capacity.
Developing dynamic capacity allocation strategies that adjust the sparsity budget based on task difficulty or inter-task similarity could extend the method to substantially longer sequences.
Additionally, while CLARE is evaluated on vision tasks with a ViT backbone, extending the capacity-aware sparse allocation principle to other modalities, architecture, and domains remains an open question.

{\small
\bibliographystyle{ieeenat_fullname}
\bibliography{egbib}

@article{li2017lwf,
  title={Learning without forgetting},
  author={Li, Zhizhong and Hoiem, Derek},
  journal={IEEE transactions on pattern analysis and machine intelligence},
  volume={40},
  number={12},
  pages={2935-2947},
  year={2017},
  publisher={IEEE}
}

@article{kirkpatrick2017EWC,
  title={Overcoming catastrophic forgetting in neural networks},
  author={Kirkpatrick, James and Pascanu, Razvan and Rabinowitz, Neil and Veness, Joel and Desjardins, Guillaume and Rusu, Andrei A and Milan, Kieran and Quan, John and Ramalho, Tiago and Grabska-Barwinska, Agnieszka and others},
  journal={Proceedings of the national academy of sciences},
  volume={114},
  number={13},
  pages={3521-3526},
  year={2017},
  publisher={National Academy of Sciences}
}

@inproceedings{marczak2024magmax,
  title={Magmax: Leveraging model merging for seamless continual learning},
  author={Marczak, Daniel and Twardowski, Bart{\l}omiej and Trzci{\'n}ski, Tomasz and Cygert, Sebastian},
  booktitle={European Conference on Computer Vision},
  pages={379-395},
  year={2024},
  organization={Springer}
}

@inproceedings{hendrycks2021ImagenetR,
  title={The many faces of robustness: A critical analysis of out-of-distribution generalization},
  author={Hendrycks, Dan and Basart, Steven and Mu, Norman and Kadavath, Saurav and Wang, Frank and Dorundo, Evan and Desai, Rahul and Zhu, Tyler and Parajuli, Samyak and Guo, Mike and others},
  booktitle={Proceedings of the IEEE/CVF international conference on computer vision},
  pages={8340-8349},
  year={2021}
}

@misc{krizhevsky2009CIFAR100,
  title={Learning multiple layers of features from tiny images.(2009)},
  author={Krizhevsky, Alex and Hinton, Geoffrey and others},
  year={2009}
}

@inproceedings{hendrycks2021ImagenetA,
  title={Natural adversarial examples},
  author={Hendrycks, Dan and Zhao, Kevin and Basart, Steven and Steinhardt, Jacob and Song, Dawn},
  booktitle={Proceedings of the IEEE/CVF conference on computer vision and pattern recognition},
  pages={15262-15271},
  year={2021}
}

@inproceedings{zhou2024EASE,
  title={Expandable subspace ensemble for pre-trained model-based class-incremental learning},
  author={Zhou, Da-Wei and Sun, Hai-Long and Ye, Han-Jia and Zhan, De-Chuan},
  booktitle={Proceedings of the IEEE/CVF Conference on Computer Vision and Pattern Recognition},
  pages={23554-23564},
  year={2024}
}

@inproceedings{gao2025MoAL,
  title={Knowledge memorization and rumination for pre-trained model-based class-incremental learning},
  author={Gao, Zijian and Jia, Wangwang and Zhang, Xingxing and Zhou, Dulan and Xu, Kele and Dawei, Feng and Dou, Yong and Mao, Xinjun and Wang, Huaimin},
  booktitle={Proceedings of the Computer Vision and Pattern Recognition Conference},
  pages={20523-20533},
  year={2025}
}

@article{zhou2025APER,
  title={Revisiting class-incremental learning with pre-trained models: Generalizability and adaptivity are all you need},
  author={Zhou, Da-Wei and Cai, Zi-Wen and Ye, Han-Jia and Zhan, De-Chuan and Liu, Ziwei},
  journal={International Journal of Computer Vision},
  volume={133},
  number={3},
  pages={1012-1032},
  year={2025},
  publisher={Springer}
}

@inproceedings{Subnetwork_PGM,
  title={Probabilistic Group Mask Guided Discrete Optimization for Incremental Learning},
  author={Wan, Fengqiang and Yang, Yang},
  booktitle={Forty-second International Conference on Machine Learning},
  year={2025},
}

@inproceedings{wang2022L2P,
  title={Learning to prompt for continual learning},
  author={Wang, Zifeng and Zhang, Zizhao and Lee, Chen-Yu and Zhang, Han and Sun, Ruoxi and Ren, Xiaoqi and Su, Guolong and Perot, Vincent and Dy, Jennifer and Pfister, Tomas},
  booktitle={Proceedings of the IEEE/CVF conference on computer vision and pattern recognition},
  pages={139-149},
  year={2022}
}

@inproceedings{rebuffi2017icarl,
  title={icarl: Incremental classifier and representation learning},
  author={Rebuffi, Sylvestre-Alvise and Kolesnikov, Alexander and Sperl, Georg and Lampert, Christoph H},
  booktitle={Proceedings of the IEEE conference on Computer Vision and Pattern Recognition},
  pages={2001-2010},
  year={2017}
}

@inproceedings{radford2021CLIP,
  title={Learning transferable visual models from natural language supervision},
  author={Radford, Alec and Kim, Jong Wook and Hallacy, Chris and Ramesh, Aditya and Goh, Gabriel and Agarwal, Sandhini and Sastry, Girish and Askell, Amanda and Mishkin, Pamela and Clark, Jack and others},
  booktitle={International conference on machine learning},
  pages={8748-8763},
  year={2021},
  organization={PmLR}
}

@article{dosovitskiy2020vit,
  title={An image is worth 16x16 words: Transformers for image recognition at scale},
  author={Dosovitskiy, Alexey and Beyer, Lucas and Kolesnikov, Alexander and Weissenborn, Dirk and Zhai, Xiaohua and Unterthiner, Thomas and Dehghani, Mostafa and Minderer, Matthias and Heigold, Georg and Gelly, Sylvain and others},
  journal={arXiv preprint arXiv:2010.11929},
  year={2020}
}

@inproceedings{mallya2018packnet,
  title={Packnet: Adding multiple tasks to a single network by iterative pruning},
  author={Mallya, Arun and Lazebnik, Svetlana},
  booktitle={Proceedings of the IEEE conference on Computer Vision and Pattern Recognition},
  pages={7765-7773},
  year={2018}
}

@inproceedings{yu2024boosting,
  title={Boosting continual learning of vision-language models via mixture-of-experts adapters},
  author={Yu, Jiazuo and Zhuge, Yunzhi and Zhang, Lu and Hu, Ping and Wang, Dong and Lu, Huchuan and He, You},
  booktitle={Proceedings of the IEEE/CVF Conference on Computer Vision and Pattern Recognition},
  pages={23219-23230},
  year={2024}
}

@inproceedings{smith2023coda,
  title={Coda-prompt: Continual decomposed attention-based prompting for rehearsal-free continual learning},
  author={Smith, James Seale and Karlinsky, Leonid and Gutta, Vyshnavi and Cascante-Bonilla, Paola and Kim, Donghyun and Arbelle, Assaf and Panda, Rameswar and Feris, Rogerio and Kira, Zsolt},
  booktitle={Proceedings of the IEEE/CVF conference on computer vision and pattern recognition},
  pages={11909-11919},
  year={2023}
}

@inproceedings{farajtabar2020orthogonal,
  title={Orthogonal gradient descent for continual learning},
  author={Farajtabar, Mehrdad and Azizan, Navid and Mott, Alex and Li, Ang},
  booktitle={International conference on artificial intelligence and statistics},
  pages={3762-3773},
  year={2020},
  organization={PMLR}
}

@article{lopez2017gradient,
  title={Gradient episodic memory for continual learning},
  author={Lopez-Paz, David and Ranzato, Marc'Aurelio},
  journal={Advances in neural information processing systems},
  volume={30},
  year={2017}
}

@inproceedings{wang2025sema,
  title={Self-expansion of pre-trained models with mixture of adapters for continual learning},
  author={Wang, Huiyi and Lu, Haodong and Yao, Lina and Gong, Dong},
  booktitle={Proceedings of the Computer Vision and Pattern Recognition Conference},
  pages={10087-10098},
  year={2025}
}

@article{wen2016learning,
  title={Learning structured sparsity in deep neural networks},
  author={Wen, Wei and Wu, Chunpeng and Wang, Yandan and Chen, Yiran and Li, Hai},
  journal={Advances in neural information processing systems},
  volume={29},
  year={2016}
}

@article{louizos2017learning,
  title={Learning sparse neural networks through $ L\_0 $ regularization},
  author={Louizos, Christos and Welling, Max and Kingma, Diederik P},
  journal={International Conference on Learning Representations},
  year={2018}
}

@article{MA2019286,
title = {Transformed $L\_1$ regularization for learning sparse deep neural networks},
journal = {Neural Networks},
volume = {119},
pages = {286-298},
year = {2019},
issn = {0893-6080},
doi = {https://doi.org/10.1016/j.neunet.2019.08.015},
url = {https://www.sciencedirect.com/science/article/pii/S0893608019302321},
author = {Rongrong Ma and Jianyu Miao and Lingfeng Niu and Peng Zhang},
}

@article{wang2022sparcl,
  title={Sparcl: Sparse continual learning on the edge},
  author={Wang, Zifeng and Zhan, Zheng and Gong, Yifan and Yuan, Geng and Niu, Wei and Jian, Tong and Ren, Bin and Ioannidis, Stratis and Wang, Yanzhi and Dy, Jennifer},
  journal={Advances in Neural Information Processing Systems},
  volume={35},
  pages={20366-20380},
  year={2022}
}

@inproceedings{yildirim2024continual,
  title={Continual learning with dynamic sparse training: Exploring algorithms for effective model updates},
  author={Yildirim, Murat Onur and Gok, Elif Ceren and Sokar, Ghada and Mocanu, Decebal Constantin and Vanschoren, Joaquin},
  booktitle={Conference on parsimony and learning},
  pages={94-107},
  year={2024},
  organization={PMLR}
}

@inproceedings{liang2024inflora,
  title={Inflora: Interference-free low-rank adaptation for continual learning},
  author={Liang, Yan-Shuo and Li, Wu-Jun},
  booktitle={Proceedings of the IEEE/CVF Conference on Computer Vision and Pattern Recognition},
  pages={23638-23647},
  year={2024}
}

@inproceedings{Wu2025SDLoRA,
  title={SD-LoRA: Scalable Decoupled Low-Rank Adaptation for Class Incremental Learning},
  author={Yichen Wu and Hongming Piao and Long-Kai Huang and Renzhen Wang and Wanhua Li and Hanspeter Pfister and Deyu Meng and Kede Ma and Ying Wei},
  booktitle={International Conference on Learning Representations},
  year={2025},
  url={https://api.semanticscholar.org/CorpusID:275820710}
}

@inproceedings{lou2026CARE,
title={Scaling Continual Learning to 300+ Tasks with Bi-Level Routing Mixture-of-Experts},
author={Lou, Meng and Fu, Yunxiang and Yu, Yizhou},
booktitle={Forty-third International Conference on Machine Learning},
year={2026},
}

@inproceedings{mallya2018piggyback,
  title={Piggyback: Adapting a single network to multiple tasks by learning to mask weights},
  author={Mallya, Arun and Davis, Dillon and Lazebnik, Svetlana},
  booktitle={Proceedings of the European conference on computer vision (ECCV)},
  pages={67-82},
  year={2018}
}

@inproceedings{aljundi2017expertGate,
  title={Expert gate: Lifelong learning with a network of experts},
  author={Aljundi, Rahaf and Chakravarty, Punarjay and Tuytelaars, Tinne},
  booktitle={Proceedings of the IEEE conference on computer vision and pattern recognition},
  pages={3366-3375},
  year={2017}
}

@inproceedings{wang2022dualprompt,
  title={Dualprompt: Complementary prompting for rehearsal-free continual learning},
  author={Wang, Zifeng and Zhang, Zizhao and Ebrahimi, Sayna and Sun, Ruoxi and Zhang, Han and Lee, Chen-Yu and Ren, Xiaoqi and Su, Guolong and Perot, Vincent and Dy, Jennifer and others},
  booktitle={European conference on computer vision},
  pages={631-648},
  year={2022},
  organization={Springer}
}

@article{chen2022adaptformer,
  title={Adaptformer: Adapting vision transformers for scalable visual recognition},
  author={Chen, Shoufa and Ge, Chongjian and Tong, Zhan and Wang, Jiangliu and Song, Yibing and Wang, Jue and Luo, Ping},
  journal={Advances in Neural Information Processing Systems},
  volume={35},
  pages={16664-16678},
  year={2022}
}

@inproceedings{zhang2023slca,
  title={Slca: Slow learner with classifier alignment for continual learning on a pre-trained model},
  author={Zhang, Gengwei and Wang, Liyuan and Kang, Guoliang and Chen, Ling and Wei, Yunchao},
  booktitle={Proceedings of the IEEE/CVF International Conference on Computer Vision},
  pages={19148-19158},
  year={2023}
}

@article{goswami2023fecam,
  title={Fecam: Exploiting the heterogeneity of class distributions in exemplar-free continual learning},
  author={Goswami, Dipam and Liu, Yuyang and Twardowski, Bart{\l}omiej and Van De Weijer, Joost},
  journal={Advances in Neural Information Processing Systems},
  volume={36},
  pages={6582-6595},
  year={2023}
}

@article{he2026loda,
  title={Task-driven subspace decomposition for knowledge sharing and isolation in lora-based continual learning},
  author={He, Lingfeng and Cheng, De and Wang, Huaijie and Yang, Xi and Wang, Nannan and Gao, Xinbo},
  journal={arXiv preprint arXiv:2603.00191},
  year={2026}
}

@article{wan2025OA-adapter,
  title={Budget-Adaptive Adapter Tuning in Orthogonal Subspaces for Continual Learning in LLMs},
  author={Wan, Zhiyi and Du, Wanrou and Li, Liang and Pan, Miao and Qin, Xiaoqi},
  journal={arXiv e-prints},
  pages={arXiv--2505},
  year={2025}
}

@article{kaushik2026shared,
  title={Shared LoRA Subspaces for almost Strict Continual Learning},
  author={Kaushik, Prakhar and Vaidya, Ankit and Chaudhari, Shravan and Chellappa, Rama and Yuille, Alan},
  journal={arXiv preprint arXiv:2602.06043},
  year={2026}
}

@article{liu2026csbora,
  title={CSBoRA: A continual learning method for large language models with true orthogonality and reduced forgetting},
  author={Liu, Yuyang and Po, Lai-Man and Hung, Farrell and Wang, Zhuohan and Wu, Haoxuan and Jiang, Zeyu and Li, Kun and Xu, Xuyuan and Cheung, Kwok-Wai},
  journal={Pattern Recognition},
  volume={179},
  pages={113782},
  year={2026},
  publisher={Elsevier}
}

@inproceedings{xiong2026oplora,
  title={Oplora: Orthogonal projection lora prevents catastrophic forgetting during parameter-efficient fine-tuning},
  author={Xiong, Yifeng and Xie, Xiaohui},
  booktitle={Proceedings of the AAAI Conference on Artificial Intelligence},
  volume={40},
  number={40},
  pages={34088-34096},
  year={2026}
}

@inproceedings{fu2025segman,
  title={SegMAN: Omni-scale context modeling with state space models and local attention for semantic segmentation},
  author={Fu, Yunxiang and Lou, Meng and Yu, Yizhou},
  booktitle={2025 IEEE/CVF Conference on Computer Vision and Pattern Recognition (CVPR)},
  pages={19077-19087},
  year={2025},
  organization={IEEE}
}

@inproceedings{lou2025overlock,
  title={Overlock: An overview-first-look-closely-next convnet with context-mixing dynamic kernels},
  author={Lou, Meng and Yu, Yizhou},
  booktitle={2025 IEEE/CVF Conference on Computer Vision and Pattern Recognition (CVPR)},
  pages={128-138},
  year={2025},
  organization={IEEE}
}

@article{lou2025transxnet,
  title={TransXNet: learning both global and local dynamics with a dual dynamic token mixer for visual recognition},
  author={Lou, Meng and Zhang, Shu and Zhou, Hong-Yu and Yang, Sibei and Wu, Chuan and Yu, Yizhou},
  journal={IEEE Transactions on Neural Networks and Learning Systems},
  volume={36},
  number={6},
  pages={11534-11547},
  year={2025},
  publisher={IEEE}
}

@inproceedings{lou2025sparx,
  title={Sparx: A sparse cross-layer connection mechanism for hierarchical vision mamba and transformer networks},
  author={Lou, Meng and Fu, Yunxiang and Yu, Yizhou},
  booktitle={Proceedings of the AAAI Conference on Artificial Intelligence},
  volume={39},
  number={18},
  pages={19104-19114},
  year={2025}
}

@inproceedings{shi2026vision,
  title={Vision transformers need more than registers},
  author={Shi, Cheng and Yu, Yizhou and Yang, Sibei},
  booktitle={2026 IEEE/CVF Conference on Computer Vision and Pattern Recognition (CVPR)},
  pages={26328-26337},
  year={2026},
  organization={IEEE}
}

@inproceedings{chen2022compound,
  title={Compound domain generalization via meta-knowledge encoding},
  author={Chen, Chaoqi and Li, Jiongcheng and Han, Xiaoguang and Liu, Xiaoqing and Yu, Yizhou},
  booktitle={2022 IEEE/CVF Conference on Computer Vision and Pattern Recognition (CVPR)},
  pages={7109-7119},
  year={2022},
  organization={IEEE}
}

@inproceedings{he2019non,
  title={Non-local context encoder: Robust biomedical image segmentation against adversarial attacks},
  author={He, Xiang and Yang, Sibei and Li, Guanbin and Li, Haofeng and Chang, Huiyou and Yu, Yizhou},
  booktitle={Proceedings of the AAAI Conference on Artificial Intelligence},
  volume={33},
  number={01},
  pages={8417-8424},
  year={2019}
}

@article{fu2024dreamda,
  title={Dreamda: Generative data augmentation with diffusion models},
  author={Fu, Yunxiang and Chen, Chaoqi and Qiao, Yu and Yu, Yizhou},
  journal={arXiv preprint arXiv:2403.12803},
  year={2024}
}

@article{li2025semantic,
  title={Semantic segmentation with scale alignment and contextual information fusion for multimodal remote sensing images},
  author={Li, Jiayuan and Wang, Zhen and Xu, Nan and You, Zhuhong},
  journal={Information Fusion},
  pages={103671},
  year={2025},
  publisher={Elsevier}
}

@article{lou2026overcoming,
  title={Overcoming Catastrophic Forgetting in Visual Continual Learning with Reinforcement Fine-Tuning},
  author={Lou, Meng and Guo, Hanzhong and Chen, Linwei and Yu, Yizhou},
  journal={arXiv preprint arXiv:2605.09640},
  year={2026}
}
}
\end{document}